%% file: main.tex
\documentclass[conference]{IEEEtran}
\input{packages.tex}

\begin{document}
    \input{0.front_matter.tex}
    \input{1.intro.tex}
    \input{2.related_work.tex}

    \input{3.proposal.tex}
    \input{4.experiments.tex}
    \input{5.evaluation.tex}

    \input{6.conclusion.tex}
    \input{7.references.tex}
\end{document}

%% file: packages.tex
\IEEEoverridecommandlockouts
\usepackage{cite}
\usepackage{amsmath,amssymb,amsfonts}
\usepackage{algorithmic}
\usepackage{graphicx}
\usepackage{textcomp}
\usepackage{xcolor}
\usepackage{multirow}
\usepackage{arydshln}
\usepackage{todonotes}
\usepackage[backref, colorlinks,urlcolor=blue]{hyperref}
\usepackage{etoolbox}
\usepackage{setspace}
\usepackage[bottom]{footmisc}
\usepackage{placeins}
\usepackage{wrapfig}
\usepackage[export]{adjustbox}

\def\BibTeX{{\rm B\kern-.05em{\sc i\kern-.025em b}\kern-.08em
    T\kern-.1667em\lower.7ex\hbox{E}\kern-.125emX}}

\usepackage{url}
    
\newcommand{\etal}{\emph{et al. }}

\patchcmd{\section}{\centering}{}{}{}

%% file: 0.front_matter.tex
 
\title {Evaluating the Transferability and Adversarial Discrimination of Convolutional Neural Networks for Threat Object Detection and Classification within X-Ray Security Imagery  }

\author{\IEEEauthorblockN{Yona Falinie A. Gaus$^1$, Neelanjan Bhowmik$^1$, Samet Ak\c{c}ay$^1$, Toby P. Breckon$^{1,2}$}
\IEEEauthorblockA{Department of \{Computer Science$^1$ $|$ Engineering$^2$\}, Durham University, UK \vspace{-0.5cm}}

}

\maketitle
\vspace{-.15cm}
\begin{abstract}
X-ray imagery security screening is essential to maintaining transport security against a varying profile of threat or prohibited items. Particular interest lies in the automatic detection and classification of weapons such as firearms and knives within complex and cluttered X-ray security imagery. Here, we address this problem by exploring various end-to-end object detection Convolutional Neural Network (CNN) architectures. We evaluate several leading variants spanning the Faster R-CNN, Mask R-CNN, and RetinaNet architectures to explore the transferability of such models between varying X-ray scanners with differing imaging geometries, image resolutions and material colour profiles. Whilst the limited availability of X-ray threat imagery can pose a challenge, we employ a transfer learning approach to evaluate whether such inter-scanner generalisation may exist over a multiple class detection problem. Overall, we achieve maximal detection performance using a Faster R-CNN architecture with a ResNet$_{101}$ classification network, obtaining 0.88 and 0.86 of mean Average Precision (mAP) for a three-class and two class item from varying X-ray imaging sources. Our results exhibit a remarkable degree of generalisability in terms of cross-scanner performance (mAP: 0.87, firearm detection: 0.94 AP). In addition, we examine the inherent adversarial discriminative capability of such networks using a specifically generated adversarial dataset for firearms detection - with a variable low false positive, as low as $5\%$, this shows both the challenge and promise of such threat detection within X-ray security imagery. 
\end{abstract}


%% file: 1.intro.tex
\section{introduction} \label{sec:intro}
\begin{figure}
    \vspace{-0.05cm}
    \centering
    \includegraphics[width=\linewidth]{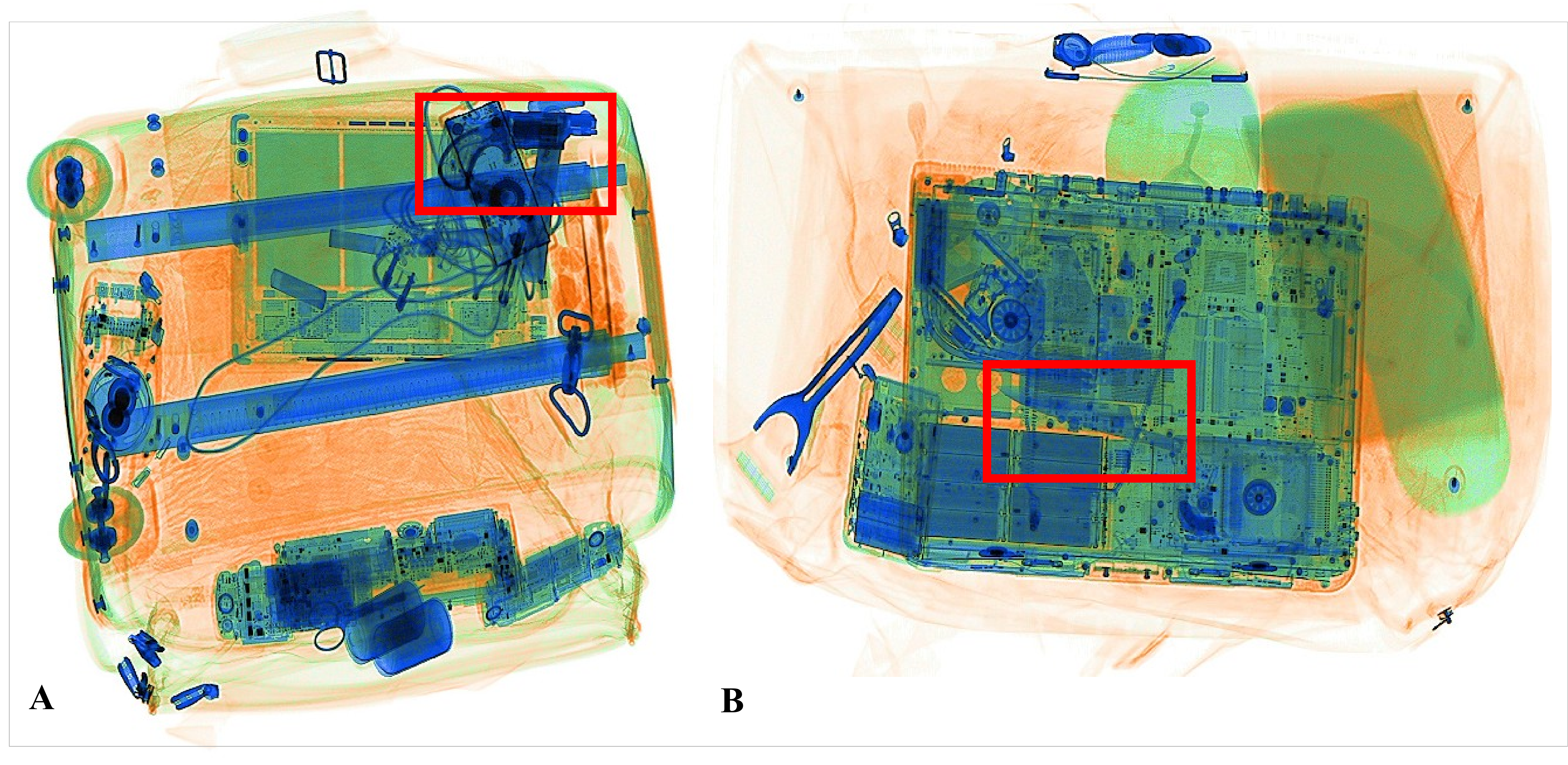}
    \vspace{-0.9cm}
    \caption{Exemplar X-ray baggage imagery with prohibited items inside (red box): (A) {\it Firearm} and (B) {\it Knife}.}
    \label{fig:sample_threat}
\end{figure}

\vspace{-.15cm}
X-ray security screening is widely used to maintain aviation, border, and transport security.
To facilitate effective screening, threat detection via scanned X-ray imagery is increasingly employed to provide a non-intrusive, internal view of scanned baggage, freight, and postal items, as illustrated in Fig. \ref{fig:sample_threat}. This produces colour-mapped X-ray images which correspond to the material properties detected via the dual-energy X-ray scanning process. Within this context, the term \textit{threat} refers to a prohibited item such as firearms, bladed weapons, or concealed explosives, etc.
\addtolength{\textfloatsep}{-0.2in}
In recent years, the rapid development of deep learning 
has brought new insight to the automation of this X-ray imagery screening task \cite{akcay2016xray, Akcay2018Xray}, where the primary task is both to localise and classify the prohibited item as it appears in the image. Therefore, in this paper, we extend the current trend of using end-to-end deep learning architectures used for this task by performing an extended evaluation of such frameworks on large-scale X-ray security imagery.
Denoted as \textit{Dbf3} \cite{Akcay2018Xray} and \textit{SIXray} \cite{Miao2019SIXray} datasets from varying X-ray scanners, we aim to provide an insight into baseline performance for several CNN architectural variants
following the work of \cite{Akcay2018Xray}.
For this study, we limit our discussion to the detection of firearms (i.e. firearms and firearms with additional parts) and sharp objects (i.e. knives) as the prohibited objects. In addition, we consider a third dataset, denoted \textit{DAD}, to specifically investigate the discriminative capability of such networks against generic \textit{adversarial} objects, manufactured to have global shape properties similar to that of a prohibited item whilst remaining benign in nature.
Subsequently, the main contributions of this paper are:
\begin{itemize}
    
    \item an exploration of three end-to-end CNN-based object detection architectures with varying network configuration for addressing prohibited item detection in X-ray imagery security, expanding the work of \cite{Akcay2018Xray, Mery2017Xrayjournal, Miao2019SIXray}.
    \item an evaluation of the inter-scanner transferability of such trained CNN models in terms of their generalizaton across varying X-ray scanner characteristics.
    \item an appraisal the trained CNN models for prohibited item discrimination against a dataset of specific adversarial objects, whose global shape characteristics closely resemble those of a firearm  within X-ray imagery.
    
\end{itemize}

%% file: 2.related_work.tex
\section{related works} \label{sec:relatedwork}
\vspace{-.15cm}
In this section, scope of the literature review here is limited to prohibited item classification and detection, presented in the following subsections:

\textit{Object Classification in X-ray Security Imagery}: 
Early work within X-ray security imagery primarily utilises handcrafted features, where a bag of visual words (BoVW) and Support Vector Machine (SVM) are applied for feature extraction and classification, respectively \cite{Bastan2013ObjectRI} \cite{kundegorski2016xray}.
Mery \etal \cite{mery2017xray} propose a method to recognise prohibited items in multiple view X-ray imagery by filtering out false positive from monocular detection performed on single views, then match it with multiple views. A BoVW approach is further employed in \cite{kundegorski2016xray} by exploring various feature point descriptors as visual word variants within a BoVW model achieving 94.0\% accuracy for two-class firearm detection with SVM classification.
The work of \cite{akcay2016xray} first introduce the use of 
CNN to address object classification task by comparing varying CNN architectures to the earlier work extensive BoVW of \cite{kundegorski2016xray}. Leveraging the use of transfer learning, \cite{akcay2016xray} shows that CNN architectures outperform BoVW features, by achieving 98.92\% detection accuracy in firearm classification.
Following \cite{akcay2016xray}, Mery \etal \cite{Mery2017Xrayjournal}  compares handcrafted features BoVW, sparse representation, codebooks with deep learning features. Consistent with the results in \cite{akcay2016xray}, deep features achieve higher results with more than 95\% accuracy in the detection of a threat. More recently, the work on \cite{Akcay2018Xray} exhaustively compares various CNN architectures to evaluate the impact of network complexity on overall performance. Fine tuning the entire network architecture for this problem domain yields 0.996\% true positive, 0.011\% false positive and 0.994\% accuracy for prohibited item detection.

\textit{Object Detection in X-ray Security Imagery}: 
Extensive experiments on object detection is conducted by Franzel \etal \cite{franzel2012xraymultiview}, where they adapt appearance-based object class detection in multiple view X-ray imagery.
Multi-view detection is shown to provide superior detection performance compared with single-view detection for handguns, with mAP of 0.645. With the recent development of object detection approach, \cite{Akcay2018Xray} examines the relative performance of traditional sliding window \cite{franzel2012xraymultiview, sermanet2013overfeat} against contemporary region-based CNN variants in X-ray security imagery \cite{Ren2015fasterr-cnn, Dai2016R_FCN, Redmon2016yolo, Girshick2015fastRCNN}. The work of \cite{Akcay2018Xray} reports the performance of a traditional sliding window driven CNN detection model based on \cite{akcay2016xray} against contemporary region-based and single forward-pass based CNN variants such as Faster R-CNN \cite{Ren2015fasterr-cnn}, R-FCN \cite{Dai2016R_FCN}, and YOLOv2 \cite{Redmon2016yolo} achieving a maximal 0.885 and 0.974 mAP over 6-class object detection and 2-class firearm detection problems respectively.
Overall, \cite{Akcay2018Xray} illustrates the real-time applicability and superiority of such integrated region based detection models within an X-ray security imagery context. Here we follow up on this theme, with our evaluation of the generalisation of such models by evaluating their inter-scanner transferability and discriminative capability against specific physical adversarial objects. 
\vspace{-0.1cm}

%% file: 3.proposal.tex
\section{proposed approach} \label{sec:proposal}
\vspace{-0.2cm}

We extend the capability of contemporary region based CNN variants by incorporating Faster R-CNN \cite{Ren2015fasterr-cnn}, Mask R-CNN \cite{he2017maskrcnn} and RetinaNet \cite{lin2017retinanet} as our prohibited item detection approach. \\
\begin{figure}
    \centering
    \includegraphics[width=\columnwidth]{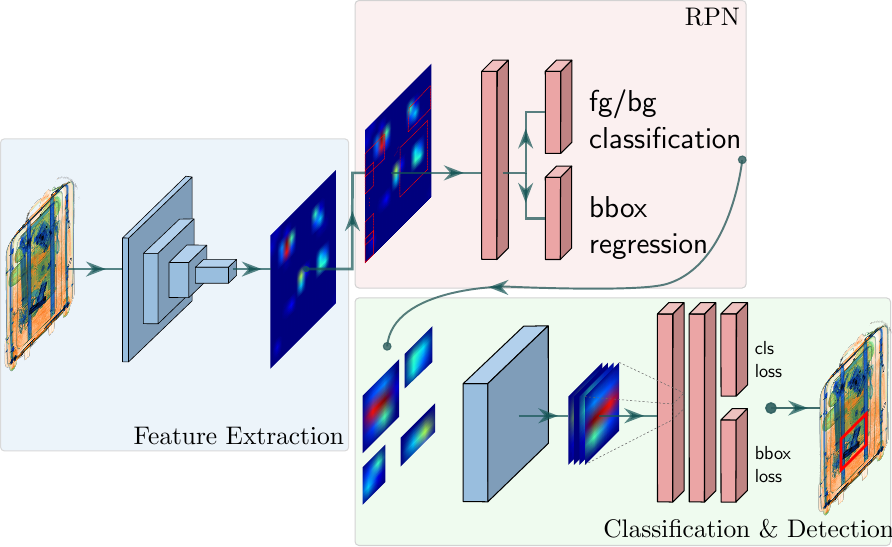}
    \vspace{-.7cm}
    \caption{Common End-to-End CNN based prohibited item detection architecture with RPN backbone.}
    \label{fig:pipeline}
\end{figure}
\textit{Faster R-CNN}: Prohibited item detection within X-ray security imagery using Faster R-CNN were first introduced in \cite{Akcay2018Xray}, where it is trained with various network architectures such as AlexNet \cite{Krizhevsky2012AlexNet}, VGG \cite{Simonyan2014VGG} and ResNet \cite{He2015ResNet}. By adding a unique region proposal network (RPN) on top of the original Fast R-CNN architecture \cite{Girshick2015fastRCNN}, it manages efficient prediction of object bounding box localization. \\
\textit{Mask R-CNN}: Following the work in \cite{Akcay2018Xray}, which builds upon Faster R-CNN in X-ray security imagery, we augment this model by adding convolutional layers to construct an object boundary segmentation mask, following the Mask R-CNN concept of \cite{he2017maskrcnn}. It is performed by adding an additional branch to Faster R-CNN that outputs an additional image mask indicating pixel membership of a given detected object. Mask R-CNN also addresses feature map misalignment, found in Faster R-CNN \cite{Ren2015fasterr-cnn} for higher resolution feature map boundaries, via bi-linear boundary interpolation. \\
\textit{RetinaNet}: RetinaNet is based on a single stage detector, which involves Focal-Loss to address class imbalance issue caused by extreme foreground-background ratio \cite{lin2017retinanet}. In terms of X-ray security imagery, the task of identifying small metal prohibited items such as a knives presents a notable challenge due to both their size characteristics and shape overlap within the general clutter of X-ray security imagery itself. Therefore, RetinaNet is considered as it offers faster processing speed and higher accuracy for small object detection, when compared to YOLO \cite{Redmon2016yolo}, thanks its unique Feature Pyramid Network (FPN) and Focal Loss function characteristics. 

Within the X-ray imagery security domain, data may be sourced from varying equipment, with different imaging parameters, X-ray energy spectra and spatial resolution  \cite{gilardoni_scanner}, \cite{nuctech_scanner}, \cite{smithsdetection_scanner}. Related work on the transferability of trained CNN models between varying X-ray scanner equipment is addressed by the work of \cite{Caldwell2017transfer}, in which they focus on transfer learning between two extremities of the X-ray screening domain in terms of scale - cargo and parcel scanning (which use very different X-ray scanner equipment due to the differences in scale). From the work of \cite{Caldwell2017transfer}, the two key issues identified in transferring knowledge across such X-ray domains are: {\it (a)} the limited availability of object of interest (prohibited item) examples, and {\it (b)} X-ray threat images appear in a different machines with very different imaging characteristcs. In this work we address a similar transferability problem between X-ray scanner equipment but within the same domain (and scale) of baggage/parcel X-ray security screening.

Our hypothesis is that a CNN model trained on a given X-ray security image dataset, gathered solely from a discrete X-ray scanner in terms of manufacturer/model, will be capable of exhibiting a high degree of generalisation in terms of performance to other such datasets gathered from varying X-ray scanner configuration (varying manufacturers/models). Here, we focus on two datasets which come from different X-ray scanners. Denoted as {\it Dbf3} \cite{Akcay2018Xray} and {\it SIXray} \cite{Miao2019SIXray}, the former dataset is from a Smith Detection X-ray scanner \cite{smithsdetection_scanner} whilst the later comes from a Nuctech scanner \cite{nuctech_scanner}. In addition, {\it Dbf3} is focused solely upon passenger carry-on baggage within an aviation security context whilst {\it SIXray} is based on security screening within a metro transit system context.
In order to investigate cross-scanner generalisation and hence transferability, our set of CNN end-to-end models are trained using on each dataset in turn and then evaluated on the other.

%% file: 4.experiments.tex
\section{experimental setup} \label{sec:exp_set}
\vspace{-.15cm}
Our experimental setup comprises of three different datasets and a common CNN training environment. As for training details, we follow same environment as in \cite{gaus2019evaluation}.

\subsection{X-ray Image Datasets} \label{ssc:db}
\begin{figure}
     \centering
     \includegraphics[width=\linewidth]{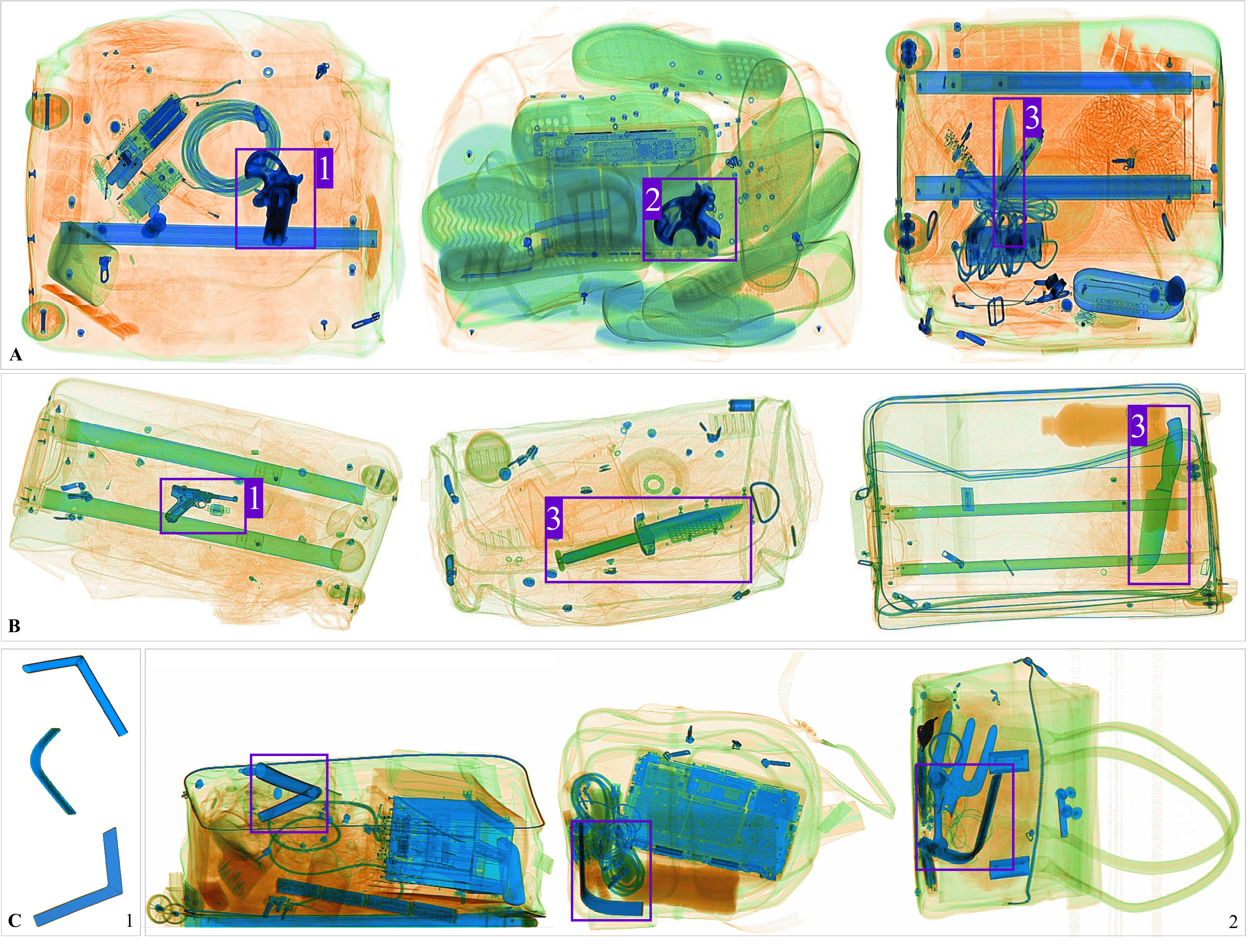}
      \vspace{-.8cm}
     \caption{Exemplar X-ray baggage images with threat objects in the purple box (1. Firearm, 2. Firearm parts, 3. Knife) from dataset (A) {\it Dbf3} and (B) {\it SIXray10}. In dataset {\it DAD} (C), the adversarial non-threat objects (C(1)), such as different shape metal objects (in purple box), are placed inside bags (C(2)).}     
    \label{fig:db_ex}
 \end{figure}

\vspace{-.15cm}
Our evaluation comprises three varying X-ray security imagery datasets: \\
{\bf Dbf3.} 
The X-ray security imagery from Durham Dataset Full Three-class ({\it Dbf3}) are generated using a Smith Detection dual-energy X-ray scanner. 
This dataset was generated using three types metallic prohibited items, where it consists of 3,192 images of {\it firearms}, 1,204 images of {\it firearms parts}, 3,207 images of {\it knives}. Each object are emplaced in a representative and varied set of test bags which cover the full dimensions of aviation cabin baggage (Fig. \ref{fig:db_ex}A). \\
{\bf SIXray10.}
We use SIXray dataset \cite{Miao2019SIXray} for prohibited item discovery in X-ray security images. It consists of $1,059,231$ X-ray images, in which six classes of $8,929$ prohibited items. These images are collected using a Nuctech dual-energy X-ray scanner, where the distribution of the general baggage/parcel items corresponds to stream-of-commerce occurrence. We use a subset of the SIXray dataset, {\it SIXray10}, which consists of five classes of prohibited items. In our experiments, we incorporate 5,083 images from two classes, 3,130 images of {\it firearms} and 1,953 images of {\it knives}, depicted in Fig. \ref{fig:db_ex}B. \\
{\bf DAD.}
Durham Adversarial Dataset ({\it DAD}) is constructed using
Gilardoni dual-energy X-ray scanner (FEP ME 640 AMX),
in the same manner as {\it Dbf3} but with artificially manufactured imitation objects, that have global shape characteristics similar to a firearm  emplaced into various baggage items. These adversarial discriminative objects are L-shaped metal objects that within X-ray imagery may resemble a firearm as depicted in Figs. \ref{fig:db_ex}C(1) $\rightarrow$ \ref{fig:db_ex}C(2). This dataset consists of 200 images of imitation (adversarial) prohibited items and 200 images of real prohibited items \{{\it firearms, knifes}\}. This dataset is created for evaluation purposes, to test the discriminative capability of the trained CNN model as to whether it can distinguish between real and imitation of prohibited items.
  

%% file: 5.evaluation.tex
\section{evaluation} \label{sec:eval}
\vspace{-.21cm}
In our evaluation, we consider two tasks related to prohibited item detection: {\it (a)} model generalisation across varying X-ray imagery datasets (Section \ref{ssec:exp_det}) and {\it (b)} model discrimination for adversarial threat objects that exhibit similar shape characteristics to the real threat objects in the same dataset. 
\subsection{Evaluation Criteria} \label{ssc:criteria}
\vspace{-.15cm}
For detection, the performance of the models is evaluated by mean average precision (mAP), as used in the seminal object detection benchmark work of \cite{lin2014coco}. We consider mean Average Precision (mAP) as our evaluation criteria following \cite{gaus2019evaluation}.
For the classification, our model performances are evaluated in terms of Accuracy (A), Precision (P), Recall (R), F-score (F1\%), True Positive (TP\%), and False Positive (FP\%) which are calculated by thresholding ($\geqslant0.7$) the intersection over union for detection.
\input{map_table.tex}

\subsection{Performance} \label{ssec:exp_det}
\vspace{-.15cm}
Performance on the prohibited item detection task is carried out by comparing
performance benchmark for these CNN models against prior work \cite{Akcay2018Xray}. Here, we present a set of intra-domain results (Section \ref{sssc:exp_det_same}, images from the same scanner dataset used for training and evaluation) against our transferablity evaluation inter-domain results  (Section \ref{sssec:exp_det_cross}, images from differing scanner datasets used for training and evaluation).  

\subsubsection{Intra-domain Results} \label{sssc:exp_det_same}

To provide reference performance measures, our CNN models are firstly trained and evaluated on the same dataset (i.e. {\it Dbf3 $\Rightarrow$ Dbf3} and {\it SIXray10 $\Rightarrow$ SIXray10}) in this set of experiments. Table \ref{Table:mAP_dbf3_sixray10} shows prohibited item  detection results for Faster R-CNN \cite{Ren2015fasterr-cnn}, Mask R-CNN \cite{he2017maskrcnn} and RetinaNet \cite{lin2017retinanet} with varying network configurations of ResNet.
We observe that the best performance (mAP = 0.88) is achieved on {\it Dbf3} by Faster R-CNN with ResNet$_{101}$ configuration, as presented in the upper part of Table \ref{Table:mAP_dbf3_sixray10}. Although Mask R-CNN and RetinaNet perform reasonably well for class {\it firearm} and {\it firearm parts}, these models perform less well on the {\it knives} class.    

The {\it firearm parts} class is absent from the {\it SIXray10} dataset (as denoted in Table \ref{Table:mAP_dbf3_sixray10}). On the remaining two classes in the {\it SIXray10} dataset, the Faster R-CNN with RestNet$_{101}$ configuration outperformed other configuration with mAP = 0.86. 
As reported in the work of \cite{Miao2019SIXray}, the highest achieved AP for the class {\it firearm}, is 90.64\% with ResNet$_{50}$. However, our model, RetinaNet with RestNet$_{101}$, produces a marginally superior AP = 0.92. The mAP results obtained for firearm detection in general are in-line with those reported in the work of \cite{Akcay2018Xray}. Overall, the class {\it knives} does not perform well and this is likely to be attributable to data imbalance in the image set used for training in addition to the greater semantic difficulty in separating this item from the background clutter.

\input{stat_table.tex}
\subsubsection{Inter-domain Results} \label{sssec:exp_det_cross}
 
\begin{figure}
     \centering
     \includegraphics[width=\linewidth]{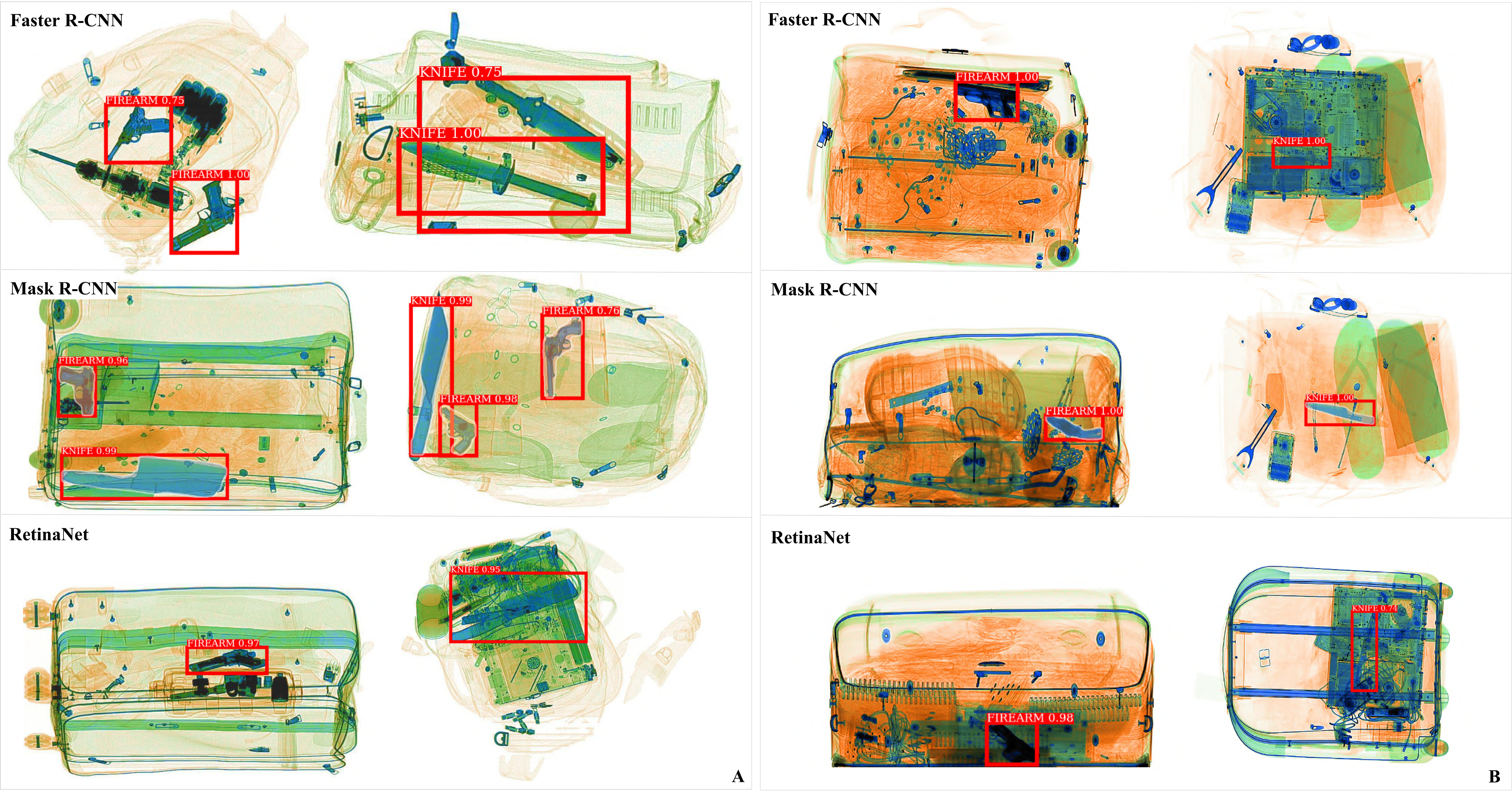}
     \vspace{-.8cm}
     \caption{Exemplar of multiple prohibited item detection for the inter-domain X-ray security training and evaluation configurations: (A) {\it Dbf3 $\Rightarrow$ SIXray10} and (B) {\it SIXray10 $\Rightarrow$ Dbf3} with varying CNN models.}     
    \label{fig:cross_det_ex}
 \end{figure}

This evaluation is to assess the CNN model performance across the X-ray security imagery from differing scanner sources.
We use the {\it Dbf3} and {\it SIXray10} datasets from varying X-ray scanners (as described in Section \ref{ssc:db}). The models are trained on one dataset and evaluated on other dataset, within which the X-Ray images are generated from a different X-ray scanner (i.e. with differing energy, geometry, resolution and colour profiles). Two sets of experiments are carried out:- firstly the models are trained using {\it Dbf3} and evaluated on the imagery from {\it SIXray10} (Table \ref{Table:mAP_cross_dbf3} - {\it Dbf3 $\Rightarrow$ SIXray10}) and secondly the inverse configuration, (i.e. {\it SIXray10} is used for training and images from {\it Dbf3} for evaluation, Table \ref{Table:mAP_cross_dbf3} - {\it SIXray10 $\Rightarrow$ Dbf3}). AP/mAP is used for performance measurement for comparison as shown in Table \ref{Table:mAP_cross_dbf3}.
We observe that for configuration {\it Dbf3 $\Rightarrow$ SIXray10}, a maximal mAP of 0.85 is achieved with Faster R-CNN, as presented in the upper part of Table \ref{Table:mAP_cross_dbf3}.
Although, RetinaNet performs equally promising for the {\it firearm} class, for the other class ({\it knives}) and globally, Faster R-CNN produces superior accuracy. In general, AP of the class {\it knives} suffers across the models due to the variation in the shape of these objects between the datasets (Figs. \ref{fig:db_ex}A/B). Detection results on the {\it SIXray10} dataset, model training performed on the {\it Dbf3} dataset ({i.e \it Dbf3 $\Rightarrow$ SIXray10}) are depicted in Fig. \ref{fig:cross_det_ex}A.  
Results for the reverse configuration, {\it SIXray10 $\Rightarrow$ Dbf3}, are presented in the lower part of Table \ref{Table:mAP_cross_dbf3} where we observe the maximally performing model is Faster R-CNN with 0.91 mAP and the best AP (= 0.94) for the {\it firearm} class. Detection results on the {\it Dbf3} dataset, whilst model training is performed on the {\it SIXray10} dataset are depicted in Fig. \ref{fig:cross_det_ex}A.
As anticipated, the {\it knives} class suffers from relatively low AP for both Mask R-CNN and RetinaNet due to the variation in visual appearance in between training and evaluation sets for this particular class (Fig. \ref{fig:cross_det_ex} A/B). In the training data, the knives are mostly placed on/under electronic items; however, the evaluation set consists of very differing shapes of knives across a diverse background.      
Overall, CNN models trained with the {\it SIXray10} dataset offers superior performance when compared to when the models are trained with {\it Dbf3} - even when evaluated on {\it Dbf3}. As a result, we can infer that although images are from differing X-ray scanners, the transferability of learnt CNN models is viable in terms of maintaining prohibited item detection performance over varying X-ray imagery sources.

\vspace{-.1cm}
\subsection{Adversarial Discriminative Objects} \label{ssec:exp_class}
\vspace{-.1cm}

Furthermore, we evaluate the discriminative capability of the CNN-based detection models we consider (Section \ref{sec:proposal}), trained for multiple class object detection (as per Table \ref{Table:mAP_dbf3_sixray10}, when tested against both real threat objects and imitation (adversarial) non-threat objects that have the same global shape and material characteristics as the real threat objects. Our test dataset for this task, {\it DAD}, is fully described in the Section \ref{ssc:db}. 
To provide an initial benchmark for performance without such adversarial examples, detection results for the three-class prohibited item problem with the {\it Dbf3} dataset and two-class threat problem within the {\it Sixray10} dataset, averaged across all object classes, are presented in Table \ref{Tab:class_intra} (calculated as per Section \ref{ssc:criteria}). Here we can observe performance such that all of the models considered consistently offer very low false positive (FP) complimented by a high true positive (TP) detection across both problems (see Table \ref{Tab:class_intra}).

To establish the impact of introducing adversarial examples, we make use of the {\it DAD} dataset (Section \ref{ssc:db}) containing our imitation (adversarial) threat-like objects, constructed as a series of simple L-shaped metal brackets, mimicking the real shape of {\it firearms}, {\it firearm parts} or {\it knives} within X-ray security imagery depending on the angle of view (see Fig. \ref{fig:db_ex}C). In order to illustrate the impact of these examples, on overall detection performance we introduce a global  {\it 'threat vs. non-threat'} detection problem on the basis that the {\it DAD} dataset (Section \ref{ssc:db}) has a $50/50$ split between X-ray security images containing a genuine {\it threat} object belonging to the set \{{\it firearm, firearm parts, knives}\} and benign ({\it non-threat}) images containing our imitation (adversarial) threat-like objects. All genuine {\it threat} and ({\it non-threat}) adversarial objects are set amongst regular benign baggage clutter. This gives rise to a simple two-class meta-problem of  {\it 'threat vs. non-threat'} by combining true positive detection for any of the set \{{\it firearm, firearm parts, knives}\} in the genuine {\it threat} object images as the class {\it threat} and conversely defining false positives as detection for any of these objects within the benign ({\it non-threat}) images that have the imitation (adversarial) threat-like objects present.

\begin{figure}
     \centering
     \includegraphics[width=0.8\linewidth]{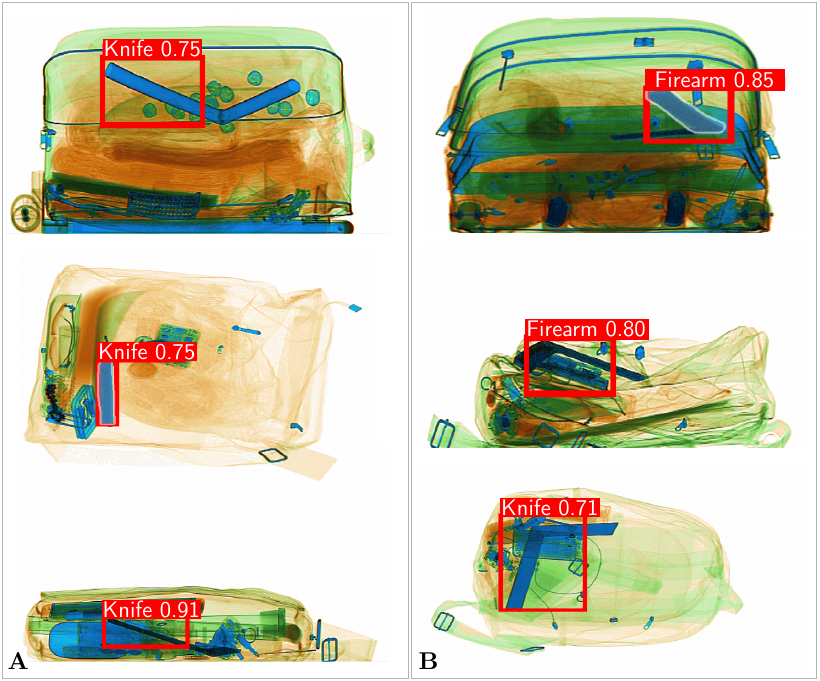}
     \vspace{-.4cm}
     \caption{Exemplar where CNN threat detection falsely detect L-shaped metal item as threat item when the models (Faster R-CNN: row.1, Mask R-CNN: row.2 and RetinaNet:  row.3) are trained on (A) {\it Dbf3} and (B) {\it SIXray10} dataset.}
    \label{fig:fake_det_ex}
 \end{figure}

As per the results of the performance benchmark shown in Table \ref{Tab:class_intra}, we evaluate the same CNN models trained on each of the {\it Dbf3} and {\it SIXray10} datasets and evaluate on {\it DAD} dataset for this two-class meta-problem, \{{\it threat, non-threat}\} (Table \ref{Tab:class}). 
With a {\it Dbf3} trained model, Faster R-CNN with ResNet$_{101}$ achieves maximal performance with the lowest FP (11.50\%) and accuracy of (84.75\%) (Table \ref{Tab:class}). However, Faster R-CNN with ResNet$_{50}$ has maximal TP (85.50\%), yet significantly higher FP. Conversely, the lowest  FP (5\%) is achieved by RetinaNet with ResNet$_{101}$ with a {\it SIXray10} trained model but this model suffers from very low TP (40.5\%). The Mask R-CNN produces 7\% FP with reasonable accuracy of 78.5\%. By comparing the performance of these models, under both standard conditions (benchmark performance in Table \ref{Tab:class_intra}) and adverserial conditions (Table \ref{Tab:class}), we can immediately see the impact of the adversarial threat-like imitators as the models get confused by the L-shaped imitation objects and wrongly classifies them as threat objects (Fig. \ref{fig:fake_det_ex}). This clearly illustrates the challenge posed by such physical adversarial object examples within achieving viable performance for automated X-ray security image classification.
\vspace{-0.2cm}

%% file: map_table.tex
\begin{table*}[ht]
\renewcommand*{\arraystretch}{0.8}
\centering
\caption{
Detection results CNN models on two datasets.
Upper: Trained and evaluated on {\it Dbf3} with three classes. Lower: Trained and evaluated on {\it SIXray10} with two classes. Class name reflects corresponding average precision (AP) for the individual object class and mAP is the mean average precision across all object classes.}
\vspace{-.3cm}
\begin{tabular}{llllcll}
\cline{1-7}
\multirow{2}{*}{\shortstack[l]{Train $\Rightarrow$ \\ Evaluation}} & \multirow{2}{*}{Model} &  \multirow{2}{*}{\shortstack[l]{Network \\ configuration}} & \multicolumn{3}{c}{Average precision} & \multirow{2}{*}{mAP} \\ \cline{4-6}
& &  & Firearm & Firearm Parts & Knives &  \\  \hline
\multirow{6}{*}{\shortstack[l]{\it Dbf3 $\Rightarrow$ \\ \it Dbf3}} & \multirow{2}{*}{Faster R-CNN \cite{Ren2015fasterr-cnn}}  & ResNet$_{50}$  & 0.87 & 0.84 & 0.76 & 0.82 \\ 
& &  ResNet$_{101}$   & {\bf 0.91}  & {\bf 0.88}  & {\bf 0.85} & {\bf 0.88} \\ \cline{2-7}

&  \multirow{2}{*}{Mask R-CNN \cite{he2017maskrcnn}} & ResNet$_{50}$ & 0.86  & 0.83  & 0.75  & 0.81  \\ 
& & ResNet$_{101}$  & 0.89  & 0.86 & 0.80 & 0.85 \\ \cline{2-7}
& \multirow{2}{*}{RetinaNet \cite{lin2017retinanet}} & ResNet$_{50}$ & 0.88 & 0.86 & 0.73 & 0.82 \\ 
& & ResNet$_{101}$  & 0.89 & 0.86 & 0.73  & 0.83 \\ \hline 
\multirow{6}{*}{\shortstack[l]{\it SIXray10 $\Rightarrow$ \\ \it SIXray10}} & \multirow{2}{*}{Faster R-CNN \cite{Ren2015fasterr-cnn}}  & ResNet$_{50}$  & 0.87 & -- & 0.77 & 0.82 \\ 
& &  ResNet$_{101}$   &  0.91  & -- & {\bf 0.81} & {\bf 0.86} \\ \cline{2-7}

& \multirow{2}{*}{Mask R-CNN \cite{he2017maskrcnn}}  & ResNet$_{50}$  & 0.87  & -- & 0.77 & 0.82 \\ 
& & ResNet$_{101}$   & 0.89 & -- & 0.79  & 0.84 \\ \cline{2-7}
& \multirow{2}{*}{RetinaNet \cite{lin2017retinanet}}  & ResNet$_{50}$  & 0.91 & -- &  0.79 & 0.85\\ 
& & ResNet$_{101}$   & {\bf 0.92}  & -- & 0.79 & 0.86 \\ \hline
\end{tabular}
\label{Table:mAP_dbf3_sixray10}
\vspace{0.2cm}
\caption{Detection results of CNN models on inter-domain datasets. Upper: Models are trained on {\it Dbf3} and evaluated on {\it SIXray10}. Lower: Models are trained on {\it SIXray10} and evaluated on {\it Dbf3}.}
\vspace{-0.3cm}
\begin{tabular}{lllcll}
\cline{1-6}
\multirow{2}{*}{\shortstack[l]{Train $\Rightarrow$ \\ Evaluation}} & \multirow{2}{*}{Model} & \multirow{2}{*}{\shortstack[l]{Network \\ configuration}} & \multicolumn{2}{c}{Average precision} & \multirow{2}{*}{mAP} \\ \cline{4-5}
 & &  & Firearm & Knives &  \\  \hline
 \multirow{3}{*}{\shortstack[l]{\it Dbf3 $\Rightarrow$ \\ \it SIXray10}} & \multirow{1}{*}{Faster R-CNN \cite{Ren2015fasterr-cnn}}  &  ResNet$_{101}$  & {\bf 0.89} & {\bf 0.80} & {\bf 0.85} \\ 

&\multirow{1}{*}{Mask R-CNN \cite{he2017maskrcnn}}  &  ResNet$_{101} $ & 0.85  & 0.77 & 0.81 \\ 
&\multirow{1}{*}{RetinaNet \cite{lin2017retinanet}} & ResNet$_{101}$ & {\bf 0.89} & 0.77  & 0.83 \\ \hline
 
 \multirow{3}{*}{\shortstack[l]{\it SIXray10 $\Rightarrow$ \\ \it Dbf3}} & \multirow{1}{*}{Faster R-CNN \cite{Ren2015fasterr-cnn}}  &  ResNet$_{101}$ & {\bf 0.94}  & {\bf 0.88}  & {\bf 0.91} \\ 

&\multirow{1}{*}{Mask R-CNN \cite{he2017maskrcnn}}  &  ResNet$_{101}$  & 0.86  & 0.72  & 0.79  \\ 
&\multirow{1}{*}{RetinaNet \cite{lin2017retinanet}} & ResNet$_{101}$ & 0.87 & 0.66 & 0.76  \\ \hline

\end{tabular}
\label{Table:mAP_cross_dbf3}
\end{table*}

%% file: stat_table.tex
\begin{table*}[ht]
\renewcommand*{\arraystretch}{0.8}
\centering
\caption{Statistical evaluation of varying CNN architectures on {\it Dbf3} and {\it SIXray10} datasets (averaged across all prohibited item  classes).}
\vspace{-0.3cm}
\begin{tabular}{lllllll|lllll}
\hline
\multirow{2}{*}{Model} & \multirow{1}{*}{Network} & \multicolumn{5}{c|}{\it Dbf3 $\Rightarrow$ Dbf3} & \multicolumn{5}{c}{\it SIXray10 $\Rightarrow$ SIXray10} \\ \cline{3-12}    
 & configuration & A & P & F1 & TP & FP & A & P & F1 & TP & FP \\ \hline
 \multirow{2}{*}{\shortstack[l]{Faster \\ R-CNN\cite{Ren2015fasterr-cnn}}} 
    & ResNet$_{50}$  &99.87 &100.00 &99.80 &99.60 &0.00 &99.07 &99.68 &99.12 &98.57 &0.36 \\ 
    & ResNet$_{101}$ & {\bf 99.96} & {\bf 100.00} & {\bf 99.93} & 99.87 & {\bf 0.00} & {\bf 99.83} &99.68 & {\bf 99.84} & {\bf 100.00} & 0.36  \\ \hline

  \multirow{2}{*}{\shortstack[l]{Mask \\ R-CNN \cite{he2017maskrcnn}}} &  ResNet$_{50}$ &99.94 &99.82 &99.91 & {\bf 100.00} &0.09 &98.65 &99.68 &98.72 &97.78 &0.36 \\ 
 &  ResNet$_{101}$ &99.93 &99.78 &99.89 &100.00 &0.11 &99.66 &99.68 &99.68 &99.68 &0.36  \\ \hline
 
 \multirow{2}{*}{RetinaNet \cite{lin2017retinanet}} 
    & ResNet$_{50}$  &97.20 &100.00 &95.62 &91.60 &0.00 &90.88 & {\bf 100.00} &90.62  &82.86  & {\bf 0.00}  \\ 
    & ResNet$_{101}$ &97.25 &100.00 &95.69 &91.74 &0.00 &90.96 &99.81 &90.74 &83.17 &0.18 \\ \hline
\end{tabular} 
\vspace{0.2cm}
\label{Tab:class_intra}
\caption{Statistical evaluation of varying CNN architecture for {\it non-threat vs threat} classification on {\it DAD} dataset.}
\vspace{-0.3cm}
\begin{tabular}{lllllll|lllll}
\hline
\multirow{2}{*}{Model} & \multirow{1}{*}{Network} & \multicolumn{5}{c|}{\it Dbf3 $\Rightarrow$ DAD} & \multicolumn{5}{c}{\it SIXray10 $\Rightarrow$ DAD} \\ \cline{3-12}    
 & configuration & A & P & F1 & TP & FP & A & P & F1 & TP & FP \\ \hline
 \multirow{2}{*}{\shortstack[l]{Faster \\ R-CNN\cite{Ren2015fasterr-cnn}}} 
    & ResNet$_{50}$  & 82.20 & 79.53 & 82.41 & {\bf 85.50} & 20.95 & {\bf 82.63} & 87.01 & {\bf 80.24} & {\bf 74.44} & 10.00 \\ 
    & ResNet$_{101}$  & {\bf 84.75} & {\bf 87.57}    & {\bf 84.16} & 81.00    & {\bf 11.50}    & 76.75    & 86.39    & 73.20    & 63.50    & 10.00 \\ \hline

  \multirow{2}{*}{\shortstack[l]{Mask \\ R-CNN \cite{he2017maskrcnn}}} &  ResNet$_{50}$ & 77.75 &    80.33 &    76.76 &    73.50 &    18.00 &    76.50 &    {\bf 89.55} &    71.86 &    60.00 &    07.00 \\ 
 &  ResNet$_{101}$ & 83.75 &    86.49 &    83.12 &    80.00 &    12.50 & 78.50 & 83.93 & 76.63 & 70.50 & 13.50 \\ \hline
 
 \multirow{2}{*}{RetinaNet \cite{lin2017retinanet}} 
    & ResNet$_{50}$ & 78.22 & 80.11 & 77.20 & 74.50 & 18.14 & 68.75 & 88.66 & 57.91 & 43.00    & 05.50  \\ 
    & ResNet$_{101}$ & 79.73    & 82.05    & 81.01    & 80.00    & 20.59 & 67.75 & 89.01 & 55.67 & 40.50    & {\bf 05.00} \\ \hline
\end{tabular}

\label{Tab:class}
\end{table*}

%% file: 6.conclusion.tex
\section{Conclusion}
\label{sec:conclusion}
\vspace{-.1cm}
This paper explores the transferability and adversarial discrimination of various end-to-end object detection Convolutional Neural Network (CNN) architectures for prohibited item detection within X-ray security imagery.
Faster R-CNN achieves superior baseline performance (0.86/0.88 mAP) over a three class prohibited item detection problem (objects: \textit{firearms, firearm parts, knives}) evaluated on two disparate datasets capturing with varying X-ray image scanner equipment. Furthermore, we directly evaluate transferability of such CNN model performance by employing cross-scanner validation to ascertain inter-scanner generalisation performance. We show that a CNN model trained on X-ray security imagery exclusively from one X-ray scanner manufacturer's device and then performance tested exclusively on separate X-ray security imagery from another manufacturer's scanner will produce strong generalisation performance despite differences in the X-ray image characteristics (0.85/0.91 mAP, two class problem - {\it firearms, knives}). This provides strong insight to the generalisation capability of the proposed method across varying X-ray imagery characteristics. Finally, we appraise the performance of such trained CNN models against physically constructed adversarial examples (imitation threat items). Whilst this shows a clear impact on generalised performance from the use of such adversarial object, it additionally illustrates the possibility of a false positive rate as low as 5\% remains under such conditions. 
\noindent{\bf Acknowledgements}: The authors would like to thank the UK Home Office for partially funding this work. Views contained within this paper are not necessarily those of the UK Home Office.
\vspace{-0.05cm}




%% file: 7.references.tex
\small{
\bibliographystyle{IEEEtran}
\bibliography{ref_mod}
}